\documentclass[preprint,12pt]{elsarticle}

\usepackage{amssymb}
\usepackage{natbib}
\usepackage{xcolor}
\usepackage{url}
\usepackage{hyperref}
\usepackage{longtable}
\usepackage{array}
\usepackage{rotating}
\usepackage{xltabular} 
\usepackage{lscape}

\journal{}

\begin{document}

\begin{frontmatter}



\title{Can Large Language Models Act as Symbolic Reasoners?}


\author[inst1]{Rob Sullivan}
\author[inst1]{Nelly Elsayed}
\affiliation[inst1]{organization={School of Information Technology},
            addressline={University of Cincinnati}, 
            city={Ohio},
            country={United States}
            }



\begin{abstract}
  The performance of Large language models (LLMs) across a broad range of domains has been impressive but have been critiqued as not being able to reason about their process and conclusions derived. This is to explain the conclusions draw, and also for determining a plan or strategy for their approach. This paper explores the current research in investigating symbolic reasoning and LLMs, and whether an LLM can inherently provide some form of reasoning or whether supporting components are necessary, and, if there is evidence for a reasoning capability, is this evident in a specific domain or is this a general capability? In addition, this paper aims to identify the current research gaps and future trends of LLM explainability, presenting a review of the literature, identifying current research into this topic and suggests areas for future work.
\end{abstract}


\begin{highlights}
\item  This paper investigates whether large language models inherently maintain reasoning abilities or if they require additional components for reasoning across different domains.
\item Investigate whether the LLM's reasoning is domain-specific or can be extended to a universal solution.
\item A comprehensive literature review of the current research gaps and future of the LLM explainability and recommendations for investigating the explainability of LLM systems.
\end{highlights}

\begin{keyword}
Large language model  \sep LLM \sep symbolic reasoning \sep explainable AI \sep responsible AI
\end{keyword}

\end{frontmatter}


\section{Introduction}\label{sec:introduction}

Newell and Simon posited that "A physical system has the necessary and sufficient means for general intelligent action" \cite{NEWELL1976}. The symbol processing approach produced many remarkable results in the early days of AI. In 2007, Nilsson predicted that "AI systems that achieve human-level intelligence will involve a combination of symbolic and non-symbolic processing" \cite{NILSSON2007}.

Large language models (LLMs), typically built on the transformer architecture \cite{vaswani2017attention}, have demonstrated impressive results in many natural language processing (NLP) applications. The underlying encoder-decoder architecture creates semantic embeddings of unstructured data, uses a self-attention mechanism as a fundamental building block and a user-generated query for generating contextual, meaningful output. The results have been impressive with many foundation models such as the Generative Pretrained Transformer (GPT) family of models providing results across a wide range of NLP use cases. Their training methodology, predicting the next word in a sequence, while effective in capturing the underlying structure of the data, is essentially encoding an understanding of the statistical nature of data and problems, rather than providing a mechanism for learning to reason \cite{zhang2022paradox}.

This criticism of LLMs, their inability to perform symbolic reasoning, has been a topic of much discussion across the AI community. Symbolic reasoning as a process of manipulating symbols to derive new symbols or to derive new symbols from existing symbols is considered a key component of general intelligence, even if constrained to specific domains \cite{NEURIPS2021_859555c7}. The question, therefore, of whether LLMs can perform symbolic reasoning is an important one, that has implications for future AI systems.

One of the common themes used to approach this problem has been through prompt engineering and augmenting using frameworks such as reAct \cite{yao2023react}, approaches such as chain-of-thought (CoT) \cite{wei2023chainofthought}, tree-of-thought (ToT) \cite{yao2023tree}, or more recently graph-of-thought \cite{Besta_2024} to provide a semblance of reasoning ability. Another common approach illustrated in the current literature is to pair an LLM with an external symbolic reasoning component. A third, more recent approach is to use a knowledge graph \cite{pan2023large}.

The advent of LLMs and their rapid application across domains has necessitated a better understanding of the processes underlying the conclusions they draw. These include whether an LLM can perform symbolic reasoning, and if so, are there restrictions on this ability, such as specific domains, or is it more general? Can LLMs perform symbolic reasoning? If so, are there restrictions on this ability? Is this restricted to specific domains, or can it be generalized across domains? The utility of LLMs and their widespread use necessitates a better understanding of the outputs generated by such systems. 

Reasoning can be thought of as a form of explanation, or as planning and strategy. These capabilities will be necessary for future systems to be able to interact with humans in a meaningful way. Being able to explain the reasoning behind a decision is a key component of trust, and to improve the confidence in a system. Mechanisms for generating strategies and planning a course of action is another key component of reasoning that will be necessary for future applications. Valmeekam et al. comment that it is often difficult to know whether LLMs are actively planning as part of their capabilities, or simply retrieving information from their knowledge base \cite{valmeekam2023planbench}. They propose a benchmark to test the planning capabilities of LLMs, examination of this is outside the scope of this paper. This memorization and retrieval, however, is a characteristic of LLMs due to the nature of their training and context windows, and could be considered a form of reasoning, as it is a form of explanation. The search performed in the scope of this paper did not include this as a core concept and instead considered explanation, to rationalize the conclusions the LLM drew from data, and planning and strategy, to consider the course of action the LLM would take in a given situation.

Through a non-exhaustive literature search, this paper aims to identify the current themes and approaches to this problem. A common theme that emerged from the literature was the use of some form of augmented prompting to guide the LLM in generating output. Kojima et al. \cite{kojima2023large}, for example, used the chain-of-thought prompting approach in a two-stage pipeline with the first focused on obtaining the steps the LLM took - a reasoning trace - and then fed this in a second time, as an additional prompt, to get the answer. Chain-of-thought prompting was commonly used\cite{Magister20231773}, \cite{wei2023chainofthought}, and \cite{yao2023tree}, the latter research using the tree-of-thought prompting approach to explore individual chain-of-thought responses using the input query and the sequence of responses built up to that point.

This suggests that the LLM itself is not capable of performing symbolic reasoning, but requires some form of external guidance. This is an interesting area for future research, as it suggests that the architecture of current LLMs, with its statistical nature, is not conducive to symbolic reasoning. This raises the question of whether future LLMs will be able to perform symbolic reasoning, or whether they will require some form of external guidance. This paper aims to provide a review of the current state of research in this area, identify the current research questions and methodologies used, and provide a discussion of the implications of the research, suggesting areas for future research.

\section{Methodology}\label{sec:methodology}

This paper used the approach proposed by Kitchenham et al. \cite{KITCHENHAM20097} and \cite{KITCHENHAM20132049} to aid in formulating our research questions and overall search and selection process. We used as sources Google Scholar, the Association for Computing Machinery (ACM) digital library, the IEEE Xplore Digital Library, and the Elsevier Scopus database for searching. 

The original set of criteria considered was to identify any paper that contained the following terms ("large language model" OR LLM) AND "symbolic reasoning" AND "symbolic representation" AND "decision-making" AND "neurosymbolic" AND "large language model" AND "symbolic reasoning". 

Several criteria were considered. First, a large language model needed to be included as a core component of the study, rather than as an ancillary tool. That is, the LLM needed to provide meaningful output for the study's use case. Passing LLM output to another component in the overall solution, pre-processing content being passed into the LLM via a form of prompt engineering, or other pre- or post-processing strategies were acceptable.

Second, a  mechanism needed to be explicitly proposed as the means for providing some form of behavior that could be considered reasoning in the context of the problem being studied. The specification for this was deliberately allowed to be flexible, as the goal was to identify a broad range of approaches to the problem. However, some form of evidentiary support that could be considered explanatory for conclusions drawn was required.

In addition to the title, author, DOI, and data source, the following data was extracted from the papers: the approach used to provide reasoning, the methodology used to prompt the LLM, the use case for the study, the architectural domain of the study, and whether the study included an internal representation of the data. 

The full text of the papers was reviewed to ensure that the inclusion criteria were met. Papers that did not meet the inclusion criteria were excluded from the study. 
\begin{table}
  \caption{Search results from libraries.}
  \centering
  \begin{tabular}{||l c c||} 
   \hline
   Library Name & Initial & Post-I/E \\ [0.5ex] 
   \hline\hline
   ACM & 4 & 0 \\ 
   IEEE Xplore  & 35 & 1 \\
   Elsevier Scopus & 16 & 7 \\
   Google Scholar  & 28 & 6 \\ [1ex] 
   \hline
  \end{tabular}

  \label{table:1}
\end{table}

\begin{landscape}
	\begin{table}
		\centering
	\caption{Data extracted from papers.}
	\label{table:2}
	\end{table}
\begin{xltabular}{0.5pt}{|p{2.5cm}| p{2.5cm}| p{2.5cm}| p{2cm} |p{2.7cm}| p{2.7cm}| p{1.5cm}|>{\arraybackslash}X}

\hline
\textbf{Author/ Date} & \textbf{DOI}  & \textbf{Approach} & \textbf{Prompt Methodology} & \textbf{Use Case} & \textbf{Architectural Domain} & \textbf{Internal Representation? }\\ 
\hline\hline
Fang et al. \cite{Fang202417985} (2024) & 10.1609/ aaai.v38i16. 29754	& Symbolic Reasoning Module & Augmented & Text-based Games & POMDP & N \\
\hline
Magister et al. \cite{Magister20231773} (2023) & 10.48550/ arXiv.2212 .08410 & Fintetuning via teacher forcing & Chain of Thought & Knowledge Distillation & Student-Teacher Models & N \\
\hline
Kojima et al. \cite{kojima2023large} (2022) & 10.48550/ arXiv.2205. 11916 & Template-based Prompting & Chain of Thought & Performance Improvement & Zero-Shot Learning & N \\
\hline
Wei et al. \cite{wei2023chainofthought} (2022) & 10.48550/ arXiv.2201. 11903 & Template-based Prompting & Chain of Thought & Performance Improvement & Foundational Models & N \\
\hline
Gaur and Saunshi \cite{Gaur20235889} (2023) & 10.48550/ arXiv.2308. 01906 & Symbolic Reasoning Module & Self-Prompting & Verification & Foundational Models & N \\
\hline
Zhang et al. \cite{zhang-etal-2023-improved} (2023) & 10.48550/ arXiv.2305. 03742  & Symbolic Reasoning Module & Baseline & Relation Extraction; Logical Reasoning & Foundational Models; Symbolic Processing & Y \\
\hline
Sheng et al. \cite{SHENG2024} (2024) & 10.1109/ ICASSP48 485.2024. 10446308 & Symbolic Reasoning Module & N/A & Logical Prediction & Foundational Models; Graph Representation; FOL & Y \\
\hline
Collins et al. \cite{collins2022structured} (2022) & 10.48550/ arXiv.2205. 05718 & Symbolic Reasoning Module & Iterative with constraints & Planning; Explanation Generation & Foundational Models (2 modes) & Y \\
\hline
Cunnington et al. \cite{CUNNINGTON2024} (2024) & 10.48550/ arXiv.2402. 01889 & Symbolic Reasoning Module; Finetuning & Baseline & Vision-Language & Foundational Models & Y \\
\hline
Xu et al. \cite{xu2024symbolllm} (2024) & 10.48550/ arXiv.2311. 09278 & Symbolic Reasoning Module; Finetuning & Baseline & Logical Construct Integration & Foundational Models; Symbolic Processing & Y \\
\hline
Sclar et al. \cite{sclar2023minding} (2023) & 10.48550/ arXiv.2306. 00924 & Symbolic Reasoning Module & Baseline & Reading Comprehension & Foundational Models; Symbolic Processing & Y \\
\hline
Yao et al. \cite{yao2023react} (2022) & 10.48550/ arXiv.2210. 03629 & Augmented prompting & Augmented & Reasoning traces; task-specific actions & Foundational Models; Few-shot learning & N \\
\hline
Shum et al. \cite{Shum202312113} (2023) & 10.18653/ v1/2023. findings- emnlp.811 & Augmented prompting &	Augmented & Prompt automation & Foundational Models & N \\
\hline
Yao et al. \cite{yao2023tree} (2023) & 10.48550/ arXiv.2305. 10601 & Augmented prompting & Tree of Thought & Exploration; lookahead & Foundational Models & N \\
\hline

\end{xltabular}

\end{landscape}

\section{Results}\label{sec:results}

The initial search returned a total of 83 papers, as shown in the 'Initial' column of Table \ref{table:1}.

An initial review of the papers returned from the search query revealed that only 14 papers were relevant to the specific research question, as shown in the post-inclusion/exclusion (Post-I/E) column of Table \ref{table:1}. This was primarily due to excluded papers referencing reasoning but not exploring the topic in any detail. The remaining papers were then reviewed in more detail to identify the specific research questions and methodologies used. The data extracted from the papers is shown in Table \ref{table:2}.

In this study, eight papers stated use of a symbolic reasoning module external to the LLM itself. Of these, two incorporated fine-tuning of the LLM in addition to using a symbolic reasoning module. One used only fine-tuning as a mechanism for providing reasoning. The remaining five papers used sophisticated prompting as an approach to guiding the LLM in generating results. This is shown in Table \ref{table:3}.

\begin{table}[h!]
  \centering
  \caption{Reasoning Approach.}
  \begin{tabular}{||l c||} 
   \hline
   Approach & Number of papers \\ [0.5ex] 
   \hline\hline
   Symbolic reasoning module & 6 \\ 
   Symbolic reasoning module and fine-tuning of LLM & 2 \\
   LLM fine-tuning only & 1 \\
   Template-based prompting & 2 \\ 
   Augmented prompting & 3 \\ [1ex] 
   \hline
  \end{tabular}

  \label{table:3}
\end{table}

Augmented prompting is defined in this study to indicate an approach above and beyond a commonly used prompting strategy such as chain-of-thought or tree-of-thought. For example, in \cite{yao2023react}, the authors introduced the concept of "thoughts" into the prompting strategy. In \cite{Shum202312113}, the prompting methodology included pruning of intermediate steps to generate rationale chains. The most commonly used methodology in the literature reviewed was chain-of-thought prompting, both in the template-based approaches and others. For example, \cite{Magister20231773} used a chain-of-thought approach within the teacher-forcing mechanism they described to train student models.

An interesting approach to the problem, using a separate reasoning model was proposed by Collins et al. \cite{collins2022structured} and included a benchmark aimed at quantifying two of the major components of reasoning: explanation and planning. Further, they contributed a parse-and-solve model that incorporated a separate reasoning model. Their results indicated that the parse-and-solve model outperformed baseline LLMs, suggesting that the statistical distribution of language, as developed within the LLM architecture, is not sufficient on its own. 

The work of Sclar et al. \cite{sclar2023minding}, which also used a separate reasoning component, took a different approach by generating belief graphs with sentences generated being fed into the LLM along with the user query. They contend that this method requires no training or fine-tuning and is flexible in light of inputs outside of the LLM's training domain. They note that their method requires a chronological ordering, which can be alleviated in other ways. The impact of this is not clear from the paper, but may be an interesting area of future research.

Xu et al. \cite{xu2024symbolllm} take the approach of generating symbol relationships from a set of external tasks, incorporate these into the LLM via a two-phase fine-tuning phase - called injection and infusion - from which a baseline LLM is generated. Their published results show some very interesting performance. It is not clear how broadly applicable the external task-set used is. This warrants further investigation.

\begin{table}[h!]
  \caption{Paper publication year.}
  \centering
  \begin{tabular}{||l c||} 
   \hline
   Publication year & Number of papers \\ [0.5ex] 
   \hline\hline
   2022 & 4 \\ 
   2023 & 6 \\
   2024 & 4 \\ [1ex] 
   \hline
  \end{tabular}

  \label{table:4}
\end{table}

As shown in Table \ref{table:4}, literature that met the criteria for this study was published between 2022 and 2024. This was hypothesized to be the case since LLMs have been built on the Transformer architecture \cite{vaswani2017attention} from 2017, the 2018 release of the Bidirectional Encoder Representations from Transformers (BERT) \cite{devlin2019bert} and the Generative Pretrained Transformer (GPT) models  released in 2018 and 2019 \cite{radford2018improving}, with the first iteration of the GPT-3 model being released in 2020. Beyond this, the chain-of-thought prompting strategy was first introduced in 2022 \cite{wei2023chainofthought}, hence the expectation that appropriate literature may be very recent.

\section{Analysis and Discussion}

This review raised the question of what we mean by reasoning. Explanation; memorization; and planning and strategy could each be considered a form of reasoning. In the literature reviewed, the term reasoning was not consistently defined. However, where some form of augmented prompting was used, the intermediate explanations would be passed into the next step of the process as well as being used to explain the reasoning to date \cite{kojima2023large}, \cite{yao2023tree}. Such approaches suggest that any reasoning being performed benefits from external guidance of some form. The work of \cite{huang2024large} confirms this and may be a potential area for future research to explore and better understand the implications and benefits of this approach.

In addition to the meaning of the term, which could be highlighted as an issue, the search terms used could be challenged. Terms such as deduction, inference, may be as pertinent. Yao et al. \cite{yao2023react} used the term 'acting' in their paper's title. This highlighted that the topic does not have a consistent definition across the literature. 

The use of some form of sophisticated prompting approach has improved the results from LLMs and, in some cases has improved the quality of the output by eliminating hallucinations and other artifacts. This is a significant improvement that gives confidence in the results generated by large language models which has been another criticism of the technology, but which is not considered here. However, incorporating a symbolic reasoning component along with the LLM suggests the overall application provides some mode of symbolic reasoning. Fang et al. \cite{Fang202417985}, conclude that LLMs can act as neurosymbolic reasoners, at least in the constrained domains of text-based games with constrained prompts. However, their use of a separate symbolic module raises questions about the extent to which the LLM itself is performing the reasoning step. 

While providing remarkable results in specific problem domains, the question of whether LLMs have the inherent capacity for symbolic reasoning when used alone is not definitively answered in the literature reviewed as part of the scope of this paper, although several authors present evidence that supports explainability if not planning and strategy \cite{Fang202417985}, \cite{kojima2023large}. The inherent probabilistic engine underpinning LLMs and the training to predict statistically likely next words in a sequence may be a limiting factor in their ability to perform symbolic reasoning. However, this requires further investigation and may be an interesting area of future research. That being said, the work of Zhang et al. \cite{zhang-etal-2023-improved} has shown that LLMs can be improved in logical reasoning tasks through differentiable symbolic programming.

The use of external symbolic reasoning modules, or sophisticated prompting strategies, may be suggestive of the limitation discussed above. Sheng et al. \cite{SHENG2024} has shown that LLMs can be integrated with first-order logic, using a symbolic reasoning module to perform logical prediction tasks. This use of symbolic reasoning modules, such as in Cunnington et al. \cite{CUNNINGTON2024}, where it was shown that foundation models can be used in neurosymbolic learning and reasoning to perform vision-language tasks, suggests another future path that may have more general-purpose utility.

Using more sophisticated prompting strategies could be argued to be a form of symbolic reasoning, as it is a form of explanation. How much further value could prompting strategies provide? Can such prompting strategies provide the external guidance and improve LLM results? Can these provide robust, transparent explanations? An area for future research may be to explore the extent to which these prompting strategies can be used to generate explanations, and to what extent, if any, they can be used to generate strategies and plans. The work of Yao et al. \cite{yao2023react} offers an interesting approach with the insertion of "thoughts" into the prompt sequence. The work of Gaur et al. \cite{Gaur20235889} has shown that LLMs can be used to perform symbolic math word problems. This approach uses a symbolic reasoning module to perform verification tasks. An interesting comment they made was how self-prompting could be used to generate better explanations, although this wasn't explored further at the time of writing.

Magister et al. \cite{Magister20231773} took an interesting approach of using smaller language models to perform reasoning tasks via a teacher-student model through knowledge distillation from the larger model to the smaller. The smaller models being taught to reason by a larger teacher model, using chain-of-thought prompting and performing knowledge distillation tasks whereby a smaller model performs the same task as the larger model. While there is evidence for improved performance, the question of whether the smaller model is performing reasoning or simply memorizing the output of the larger model is not clear.

Recent iterations of foundation models have included additional volumes of data specific to activities such as programming. This has allowed for the development of models that can perform symbolic reasoning tasks in the context of programming, such as Meta's Code Llama model \cite{roziere2024code}. An interesting question for future research would be the extent to which, for example, such an LLM could generate an intermediate symbolic representation or could be used to generate a symbolic representation of a program.

The question of interpretability, within the broader area of explainable AI (XAI), is seeing a significant amount of research interest. During the review phase, new papers pertinent to this research question were published. Of particular relevance were Lightman et al. \cite{lightman2023letsverifystepstep}, Srivastava et al. \cite{srivastava2024functionalbenchmarksrobustevaluation}, and Mirzadeh et al. \cite {mirzadeh2024gsmsymbolicunderstandinglimitationsmathematical} were included in this paper as they provided significant insights into particular parts of the problem domain. 

Mirzadeh et al hypothesize that LLMs are not capable of genuine logical reasoning. They further conclude that any reasoning ascribed to such models is an attempt to replicate reasoning associated with their training data.  Intuitively, it would seem unlikely that an LLM could exhibit reasoning on its own as a fundamental part of the architectures of such models is to predict the next token in the sequence based upon statistical distributions. However, this is still an open question. Srivasta et al propose an evaluation framework using \emph{functional variants} of benchmarks and uncovered a \emph{reasoning gap} between state-of-the-art models. They issue a caveat that these gaps may be shrunk using sophisticated prompting strategies. As indicated above, such prompting approaches do seem to positively improve the results from LLMs and it suggests this would be an interesting area of future research to determine how different prompting strategies would affect the justification parameters for outcomes generated by LLMs.  

Lightman et al.'s work, and the closely-related work of Uesato et al. \cite{uesato2022solvingmathwordproblems} described two approaches to interpretation and verification; one based on the \emph{outcome} itself, the other on the \emph{process}. Using outcome-supervised reward models (ORMs) and process-supervised reward models (PRMs), their work focused on training reward models as a mechanism for improving the performance of fine-tuned LLMs. They conclude that the PRM-based approach not only enables better results fro the LLMs, but that this approach also provides a more interpretable output from the chain-of-thought strategy since it provides feedback on each step in the chain. 
\color{black}

\section{Conclusion}

Current literature is divided on the question of whether large language models have an inherent capability to perform symbolic reasoning. Fang et al. \cite{Fang202417985} concluded that LLMs can act as neurosymbolic reasoners, at least in the constrained domains of text-based games with constrained prompts. However, they used a separate symbolic module, so further investigation would be needed to determine if the LLM itself was performing the reasoning step or not. Zhang et al. \cite{zhang2022paradox} concluded that the model they studied - the BERT model - had not learned a reasoning function but instead learned statistical patterns. Sophisticated prompting strategies have elicited evidence of explainability in the results generated by LLMs. Constraining the definition of reasoning in this way suggests that the external guidance provided by these prompting strategies is necessary for the LLM to generate meaningful output. Aligning LLMs with symbolic reasoning modules such as in Fang et al. \cite{Fang202417985}, Cunnington et al. \cite{CUNNINGTON2024}, and others suggests the need for additional components, possibly other language models as in the work of Magister et al. \cite{Magister20231773}, to provide the reasoning capability. The conclusion reached from the literature review performed is that the probabilistic architecture of current LLMs, is not able to inherently perform symbolic reasoning, requiring external impetus to move in this direction. While progress is evident in the explainability dimension, planning and strategy appear much less mature. 

Process-supervised reward model (PRM) research suggests a novel path toward providing granular interpretation of conclusions reached by LLMs. In domains such as healthcare and finance, this could generate the insights into the decision-making of the LLM and thus support LLM adoption in problem areas within these domains. One question that this raises, however, relates to a characteristic of the experiments undertaken as part of current research. The problem areas studied allowed for many solutions to the same problem. While this can be seen in mathematical problems, it is not clear that this is the case in, for example, medical diagnosis. If there are significantly fewer solution paths, the implications for the PRM-based approach is not currently understood and would be an interesting area for future research.  
\color{black}
 \bibliographystyle{elsarticle-num-names} 
 \bibliography{cas-refs}





\end{document}